\documentclass[sigconf,screen]{acmart}
\usepackage{array}
\usepackage{booktabs} 
\usepackage{epsfig}
\usepackage{graphicx}
\usepackage{amsmath}
\usepackage{amssymb}
\usepackage{color}
\usepackage{subfig}
\usepackage{comment}
\usepackage{enumerate}
\usepackage{arydshln}
\usepackage{rotating}
\usepackage{adjustbox}
\usepackage{hyperref}
\usepackage{amsfonts}
\usepackage{amsmath}
\usepackage{url}
\usepackage{caption} 

\newcolumntype{H}{>{\setbox0=\hbox\bgroup}c<{\egroup}@{}}
\newcommand*{\EmbeddingRoot}{.}
\renewcommand{\comment}[1]{}
\newcommand{\bv}[1]{\mathbf{#1}}
\newcommand{\T}[1]{\mathrm{T}}
\def\eg{\textit{e.g. }}
\def\wrt{\textit{w.r.t.} }
\def\etal{\textit{et al.} }
\def\cf{\emph{c.f}. }
\def\etc{\textit{etc.} }
\def\ie{\textit{i.e. }}

\setcopyright{acmlicensed}
\copyrightyear{2017}
\acmYear{2017}
\acmConference{MM'17}{}{October 23--27, 2017, Mountain View, CA, USA.}
\acmPrice{15.00}
\acmDOI{https://doi.org/10.1145/3123266.3123297}
\acmISBN{ISBN 978-1-4503-4906-2/17/10}

\begin{document}
\title{Query-adaptive Video Summarization via Quality-aware Relevance Estimation\vspace{-1.5ex}}

\author{Arun Balajee Vasudevan\footnotemark[1]\footnotemark[2], Michael Gygli\footnotemark[1]\footnotemark[2]\footnotemark[3], Anna Volokitin\footnotemark[2], Luc Van Gool\footnotemark[2]\footnotemark[4]\vspace{-1.2ex}}

\affiliation{%
  \institution{\footnotemark[2]ETH Zurich \hspace{1cm} \footnotemark[4] KU Leuven\hspace{1cm} \footnotemark[3] Gifs.com}
\vspace{0.5ex}}
\email{ {arunv,gygli,anna.volokitin,vangool}@vision.ee.ethz.ch }

\begin{abstract}
Although the problem of automatic video summarization has recently received a lot of attention, the problem of creating a video summary that also highlights elements relevant to a search query has been less studied. 
We address this problem by posing query-relevant summarization as a video frame subset selection problem, which lets us optimise for summaries which are simultaneously diverse, representative of the entire video, and relevant to a text query.
We quantify relevance by measuring the distance between frames and queries in a common textual-visual semantic embedding space induced by a neural network. In addition, we extend the model to capture query-independent properties, such as frame quality. We compare our method against previous state of the art on textual-visual embeddings for thumbnail selection and show that our model outperforms them on relevance prediction.  
Furthermore, we introduce a new dataset, annotated with diversity and query-specific relevance labels. On this dataset, we train and test our complete model for video summarization and show that it outperforms standard baselines such as Maximal Marginal Relevance.
\end{abstract}

\maketitle

\footnotetext[1]{Authors contributed equally}

\section{Introduction}
Video recording devices have become omnipresent. 
Most of the videos taken with smartphones, surveillance cameras and wearable cameras are recorded with a \textit{capture first, filter later} mentality.  However, most raw videos never end up getting curated and remain too long, shaky, redundant and boring to watch.  This raises new challenges in searching both within and across videos.

The problem of making videos content more accessible has spurred research in automatic tagging~\cite{Qi2007,Ballan2014,Mazloom2016}  and video summarization~\cite{Sun2014,Gygli2015,Potapov2014,Ghosh2012,Khosla,Lu2013,Park2014,Kim,Zhao}.
In automatic tagging, the goal is to predict meta-data in form of tags, which makes videos searchable via text queries. Video summarization, on the other hand, aims at making videos more accessible by reducing them to a few interesting and representative frames~\cite{Khosla,Ghosh2012} or shots~\cite{Gygli2015,Song2015}. 

\begin{figure}[t]
\centering
\begin{tabular}{l}
\noindent
 \includegraphics[width=1.0\linewidth]{\EmbeddingRoot/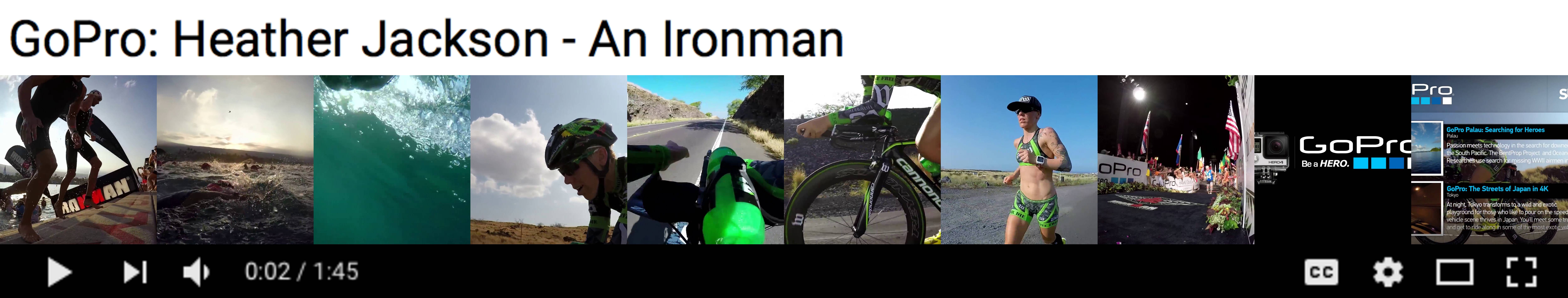}\\
\noindent 
\includegraphics[width=1.0\linewidth]{\EmbeddingRoot/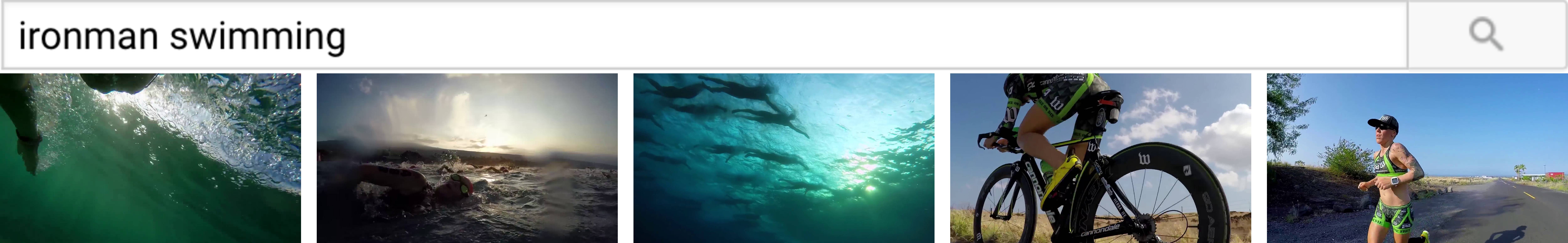}\\
\noindent 
\includegraphics[width=1.0\linewidth]{\EmbeddingRoot/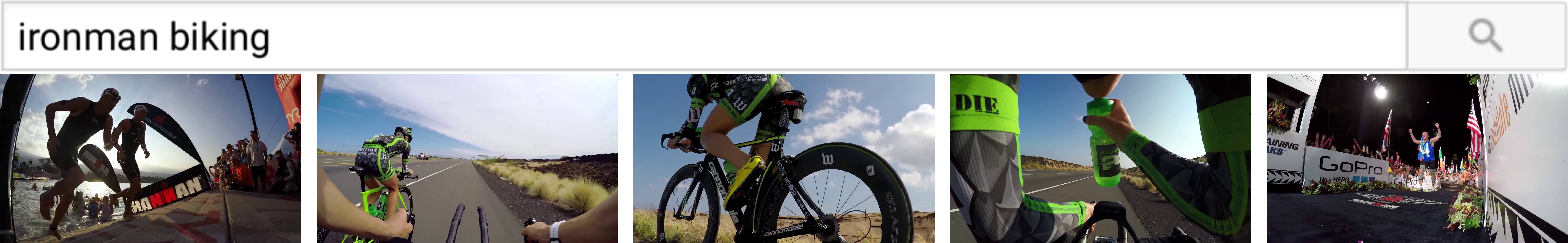}\\
\end{tabular}
\vspace{-0.3cm}
\caption{Our query-adaptive video summarization model picks frames that are relevant to the query while also giving a sense of entire video. We want to summarise a video of an \textit{ironman} competition, in which participants swim, bike and run. Query-adapted summaries are representative by showing all three sports, while placing more focus on the frames matching the query.
}
\vspace{-0.40cm}
\label{fig:teaser}
\end{figure}
This paper combines the goals of summarising videos and makes them searchable with text. Specifically, we propose a novel method that generates video summaries adapted to a text query (See Fig.~\ref{fig:teaser}).
Our approach improves previous works in the area of textual-visual embeddings~\cite{Kiros2014,Liu2015} and proposes an extension of an existing video summarization method using submodular mixtures~\cite{Gygli2015} for creating summaries that are query-adaptive.

\begin{figure*}[t]
\centering
\adjustbox{trim={.08\width} {.295\height} {0.02\width} {.325\height},clip}
    {\includegraphics[width=0.95\linewidth]{\EmbeddingRoot/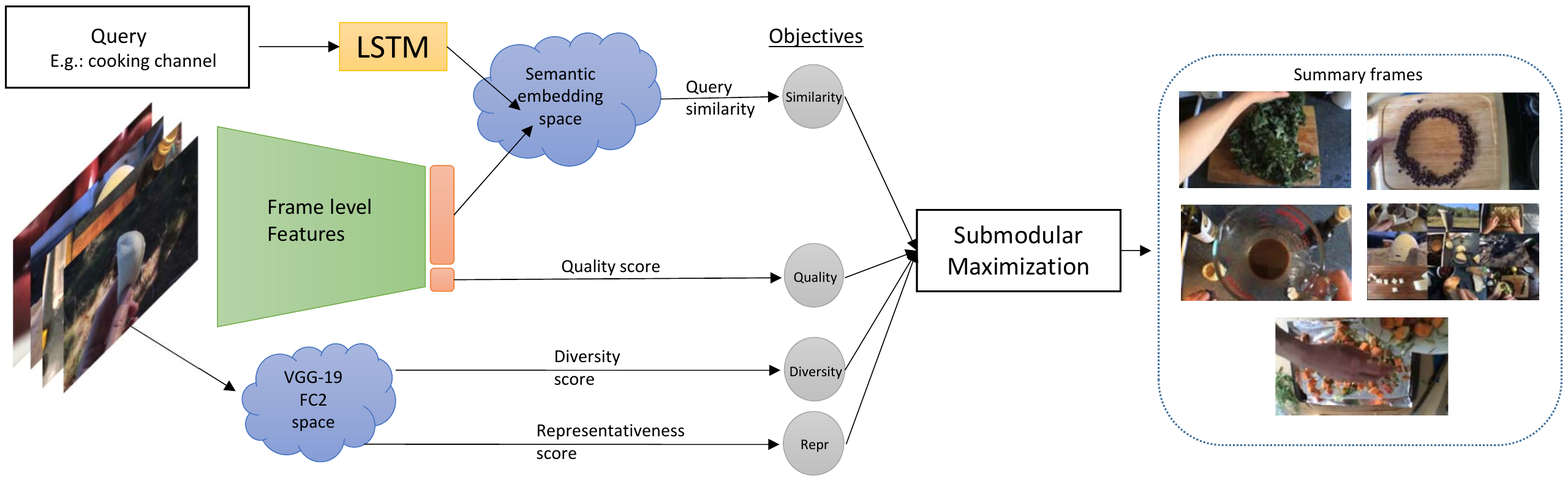}}
    \vspace{-0.27cm}
 \caption{Overview of our approach. We show how a summary is created from an example video and the query \textit{Cooking channel}. We obtain a query adaptive summary by selecting a set of keyframes from the video using our quality-aware relevance model and submodular mixtures, as explained in Sec.~\ref{sec:relevance} and ~\ref{sec:summarization}.}
 \label{fig:embedding}
 \vspace{-0.25cm}
 \end{figure*}

Our method for creating query-relevant summaries consists of two parts.  We first develop a relevance model which allows us to rank frames of a video according to their relevance given a text query. Relevance is computed as the sum of the cosine similarity between embeddings of frames and text queries in a learned visual-semantic embedding space and a query-independent term. While the embedding captures semantic similarity between video frames and text queries, the query-independent term predicts relevance based on the quality, composition and the interestingness of the content itself.
We train this model on a large dataset of image search data~\cite{hua2013clickture} and our newly introduced Relevance and Diversity dataset (Section~\ref{sec:rad}).
The second part of the summarization system is a framework for optimising the selected set of frames not only for relevance, but also for representativeness and diversity using a submodular mixture of objectives. Figure~\ref{fig:embedding} shows an overview of our complete pipeline.
We publish our codes and demos \footnote[2]{\url{https://github.com/arunbalajeev/query-video-summary}} and make the following contributions:
\begin{itemize}
\item Several improvements on learning a textual-visual embedding for thumbnail selection compared to the work by Liu~\etal ~\cite{Liu2015}. These include better alignment of the learning objective to the task at test time and modeling the text queries using LSTMs, fetching significant performance gains.
\vspace{3pt}
\item A way to model semantic similarity and quality aspects of frames jointly, leading to better performance compared to using the similarity to text queries only.

\vspace{3pt}
\item We adapt the submodular mixtures model for video summarization by Gygli~\etal~\cite{Gygli2015} to create query-adaptive and diverse summaries using our frame-based relevance model.
\vspace{-5pt}
\item A new video thumbnail dataset providing query relevance and diversity labels. 
As the judgements are subjective, we collect multiple annotations per video and analyse the consistency of the obtained labelling.
\end{itemize}

\section{Related Work}
 
The goal of video summarization is to select a subset of frames that gives a user an idea of the video's content at a glance~\cite{Truong2007}.
To find informative frames for this task, two dominant approaches exist: (i) modelling generic frame interestingness~\cite{Ghosh2012,Gygli2016} or (ii) using additional information such as the video title or a text query to find relevant frames~\cite{Liu2009,Song2015,Liu2015}.
In this work we combine the two into one model and make several contributions for query-adaptive relevance prediction. Such models are related to
automatic tagging~\cite{Qi2007,Ballan2014,Mazloom2016}, 
textual-visual embeddings~\cite{Frome2013,Socher2014,Liu2015}
and image description~\cite{Das2013,Barbu2012,Karpathy2015,Donahue2015b,Mao2014,Chen2014,Karpathy2014c,Fang2015} .
In the following we discuss approaches for video summarization, generic interestingness prediction models and previous works for obtaining embeddings.

\vspace{-0.1cm}
\bigskip
\noindent
\textbf{Video summarization.}
Video summarization methods can be broadly classified into abstractive and extractive approaches.  
Abstractive or compositional approaches transform the initial video into a more compact and appealing representation, e.g. hyperlapses~\cite{Kopf2014}, montages~\cite{Sun2014a} or video synopses~\cite{Pritch2008}. The goal of extractive methods is instead to select an informative subset of keyframes~\cite{Wolf1996,Ghosh2012,Khosla,Kim} or video segments~\cite{Gygli2015,Lu2013} from the initial video.
Our method is extractive. Extractive methods need to optimise at least two properties of the summary: the quality of the selected frames and their diversity~\cite{Sharghi2016,Gygli2015,Gong2014}. Sometimes, additional objectives such as temporal uniformity~\cite{Gygli2015} and relevance~\cite{Sharghi2016} are also optimised.
The simplest approach to obtain a representative and diverse summary is to cluster videos into events and select the best frame per event~\cite{DeAvila2011}. More sophisticated approaches jointly optimise for importance and diversity by using determinantal point process (DPPs)~\cite{Gong2014,Sharghi2016,zhang2016video} or submodular mixtures~\cite{Lin,Gygli2015}.
Most related to our paper is the work of Sharghi~\etal~\cite{Sharghi2016}, who present an approach for query-adaptive video summarization using DPPs.
Their method however limits to a small, fixed set of concepts such as \textit{car} or \textit{flower}. The authors leave handling of unconstrained queries, as in our approach, for future work.
In this work, we formulate video summarization as a maximisation problem over a set of submodular functions, following~\cite{Gygli2015}.

\vspace{-0.1cm}
\bigskip
\noindent
\textbf{Frame quality/interestingness.}
Most methods that predict frame interestingness are based on supervised learning.
The prediction problem can be formulated as a classification~\cite{Potapov2014}, regression~\cite{Ghosh2012,zen2016mouse}, or, as is now most common, as a ranking problem~\cite{Sun2014,Gygli2016,yao2016highlight,sun2017semantic}.
To simplify the task, some approaches assume the domain of the video given and train a model for each domain~\cite{Potapov2014,Sun2014,yao2016highlight}.

An alternative approach based on unsupervised learning, proposed by Xiong~\etal~\cite{xiong2014detecting}, detects ``snap points'' by using a web image prior. Their model considers frames suitable as keyframes if the composition of the frames matches the composition of the web images, regardless of the frame content. Our approach is partially inspired by this work in that it predicts relevance even in the absence of a query, but relies on supervised learning.

\vspace{-0.1cm}
\bigskip
\noindent
\textbf{Unconstrained Textual-visual models.}
Several methods exist that can retrieve images given unconstrained text or vice versa~\cite{Frome2013,Mao2014,Karpathy2014c,Karpathy2015,Donahue2015b,Fang2015,habibian2016videostory}.
These typically project both modalities into a joint embedding space~\cite{Frome2013}, where semantic similarity can be compared using a measure like cosine similarity. Word2vec~\cite{Mikolov2013a} and GloVe~\cite{pennington2014glove} are popular choices to obtain the embeddings of text.
Deep image features are then mapped to the same space via a learned projection.  Once both modalities are in the same space, they may be easily compared~\cite{Frome2013}. A multi-modal semantic embedding space is often used by Zero-shot learning approaches~\cite{Frome2013, norouzi2013zero,jain2015objects2action} to predict test labels which are unseen in the training. Habibian~\etal~\cite{habibian2016videostory}, in the same spirit, propose zero-shot recognition of events in videos by learning a video representation that aligns text, audio and video features.
Similarly, Liu~\etal~\cite{Liu2015} use textual-visual embeddings for video thumbnail selection. 
Our relevance model is based on Liu~\etal~\cite{Liu2015}, but we provide several important improvements. (i) Rather than keeping the word representation fixed, we jointly optimise the word and image projection. (ii) Instead of embedding each word separately, we train an LSTM model that combines a complete query into one single embedding vector, thus it even learns multi-word combinations such as \textit{visit to lake} and \textit{Star Wars movie}.
(iii) In contrast to Liu~\etal~\cite{Liu2015}, we directly optimise the target objective.  
Our experiments show that these changes lead to significantly better performance in predicting relevant thumbnails.

\section{Method for Relevance Prediction}
\label{sec:relevance}

The goal of this work is to introduce a method to automatically select a set of video thumbnails that are both relevant with respect to a query, but also diverse enough to represent the video.
To later optimise relevance and diversity jointly, we first need a way to evaluate the relevance of frames.  

Our relevance model learns a projection of video frames $v$ and text queries $t$ into the same embedding space. We denote the projection of $t$ and $v$ as $\mathbf{t}$ and $\mathbf{v}$, respectively. Once trained, the relevance of a frame $v$ given a query $t$ can be estimated via some similarity measure. As~\cite{Frome2013}, we use the cosine similarity
\begin{equation} \label{eq:similarity}
s(\mathbf{t},\mathbf{v}) = \frac {\mathbf {t} \cdot \mathbf {v} } { \|\mathbf {t} \|\|\mathbf {v} \|}.
\end{equation}
While this lets us assess the semantic relevance of a frame \wrt a query, it is also possible to make a prediction on the suitability as thumbnails \textit{a priori}, based on the frame quality, composition, \etc~\cite{xiong2014detecting}.
Thus, we propose to extend above notion of relevance and model the quality aspects of thumbnails explicitly by computing the final relevance as the sum of the embedding similarity and the query-independent frame quality term,~\ie
\begin{equation} \label{eq:relevance}
r(t,v) = s(\mathbf{t},\mathbf{v}) + q_v,
\end{equation}
where $q_v$ is a query-independent score determining the suitability of $v$ as a thumbnail, based on the quality of a frame.
 
In the following, we investigate how to formulate the task of obtaining the embeddings $\mathbf{t}$ and $\mathbf{v}$, as well as $q_v$.

\subsection{Training objective}
Intuitively, our model should be able to answer ``What is the best thumbnail for this query?''. Thus, the problem of picking the best thumbnail for a video is naturally formulated as a ranking problem. We desire that the embedding vectors of a query and frame that are a good match are more similar than ones of the same query and a non-relevant frame\footnote{Liu~\etal~\cite{Liu2015} does the inverse. It poses the problem as learning to assign a higher similarity to  corresponding frame and query than to the same frame and a random query. Thus, the model learns to answer the question ``what is a good query for this image?''.
}.
Thus, our model should learn to satisfy the rank constraint that given a query $t$,  the relevance score of the relevant frame $v^+$ is higher than the relevance score of the irrelevant frame $v^-$:
\begin{equation} \label{eq:obj1}
r(t,v^+) > r(t,v^-).
\end{equation}
Alternatively, we can train the model by requiring that both the similarity score and the quality score of the relevant frame are higher than for the irrelevant frame explicitly, rather than imposing a constraint only on their sum, as above.  In this case we would be imposing the two following constraints:
\begin{equation}\label{eq:obj2}
\begin{aligned} 
s(\mathbf{t},\mathbf{v^+}) &> s(\mathbf{t},\mathbf{v^-}) \\
q_{v^+} &> q_{v^-}.
\end{aligned}
\end{equation} 
Experimentally, we find that training with these explicit constraints leads to slightly improved performance (See Tab.~\ref{tab:detailed-expts}).

In order to impose these constraints and train the model, we define the loss as
\begin{equation}\label{eq:loss}
\begin{aligned} 
loss(t, v^+, v^-) &= l_p \left(\max\left(0, \gamma - s( \mathbf {t}, \mathbf {v^+}) + s(\mathbf {t},\mathbf {v^-})\right)\right) \\
&+ l_p \left(\max\left(0, \gamma - q_{v^+} + q_{v^-}\right)\right),
\end{aligned}
\end{equation}
where $l_p$ is a cost function 
and $\gamma$ is a margin parameter.
We follow~\cite{Gygli2016} and use a Huber loss for $l_p$, \ie the robust version of an $l_2$ loss.
Next, we describe how to parametrize the $\mathbf{t}$, $\mathbf{v}$ and $q_v$, so that they can be learned.

\subsection{Text and Frame Representation}
\label{sec:textrep}
 We use a convolutional neural network (CNN) for predicting $\mathbf{v}$ and $q_v$, while $\mathbf{t}$ is obtained via a recurrent neural network.
To jointly learn the parameters of these networks, we use a Siamese ranking network, trained with triplets of $(t, v^+, v^-)$ where the weights for the subnets predicting $v^+$ and $v^-$ are shared. We provide the model architecture in supplementary material. 
We now describe the textual representation $\mathbf{t}$ and the image representations $\mathbf{v}$ and $q_v$ in more detail.

\bigskip
\noindent
\textbf{Textual representation.} As a feature representation $\mathbf{t}$ of the textual query $t$, we first project each word of the query into a $300$-dimensional semantic space using the word2vec model~\cite{mikolov2013distributed}, which is trained on GoogleNews dataset. We fine-tune the word2vec model using the unique queries from the Bing Clickture dataset~\cite{hua2013clickture} as sentences.
Then, we encode the individual word representations into a single fixed-length embedding using an LSTM~\cite{hochreiter1997long}. We use a many-to-one prediction, where the model outputs a fixed length output at the final time-step.
This allows us to emphasize visually informative words and handle phrases.

\bigskip
\noindent
\textbf{Image representation.} To represent the image, we leverage the feature representations of a pre-trained VGG-19 network~\cite{simonyan2014very} on ImageNet~\cite{deng2009imagenet}.
We replace the softmax layer(1000 nodes) of VGG-19 network with a linear layer $M$ with 301 dimensions. The first 300 dimensions are used as the embedding $\mathbf{v}$, while the last dimension represents the quality score $q_v$.

\section{Summarization model}
\label{sec:summarization}

We use the framework of submodular optimization to create summaries that take into account multiple objectives~\cite{Lin}.
In this framework, summarization is posed as the problem of selecting a subset (in our case, of frames) $\bv{y^*}$ that maximizes a linear combination of submodular objective functions $\bv{f}(\bv{x_\mathcal{V},y})=[f_1(\bv{x_\mathcal{V},y}),...,f_n(\bv{x_\mathcal{V},y})]^T$. Specifically,
\begin{equation}
\bv{y^*} = \arg\max_{\bv{y}\in\mathcal{Y_V}}\bv{w^\T{}}\bv{f}(\bv{x_\mathcal{V},y}),
\label{eq:inference}
\end{equation}
where $\mathcal{Y_V}$ denote the set of all possible solutions $\bv{y}$ and $\bv{x_\mathcal{V}}$ the features of video $\mathcal{V}$.
In this work, we assume that the cardinality $|\bv{y}|$ is fixed to some value $k$ (we use $k=5$ in our experiments).

For non-negative weights $\bv{w}$, the objective in Eq.~\eqref{eq:inference} is submodular~\cite{Krause2011}, meaning that it can be optimized near-optimally in an efficient way using a greedy algorithm with lazy evaluations~\cite{Nemhauser1978,Minoux1978}. 

\bigskip
\noindent
\textbf{Objective functions.}
We choose a small set of objective functions, each capturing different aspects of the summary.

\begin{enumerate}
\vspace{3pt}
\item Query similarity $\bv{f}(\cdot,\cdot) =  \sum_{v \in \bv{y}} s(\mathbf{t},\mathbf{v})$ where $\mathbf{t}$ is the query embedding, $\mathbf{v}$ is frame embedding and $s(\cdot,\cdot)$ denotes the cosine similarity defined in Eq.~\eqref{eq:similarity}.
\vspace{3pt}
\item Quality score $\bv{f}(\cdot,\cdot) =  \sum_{v \in \bv{y}} q_{v}$, where $q_{v}$ represents score that is based on the quality of $v$ as a thumbnail. 
This model scores the image relevance in a query-independent manner based on properties such as contrast, composition, etc. 

\vspace{3pt}
\item Diversity of the elements in the summary \\
$\bv{f}(\bv{x_\mathcal{V},y}) =  \sum_{i\in \bf{y}}\min\limits_{j< i} D_{x_\mathcal{V}}(i,j)$,
according to some dissimilarity measure $D$. We use the Euclidean distance in of the FC2 features of the VGG-19 network for $D$\footnote{Derivation of submodularity of this objective is provided in the suppl.}.
\vspace{3pt}
\item Representativeness~\cite{Gygli2015}. This objective favors selecting the medoid frames of a video, such that the visually frequent frames in the video are represented in the summary.
\end{enumerate}
\smallskip
\noindent
\textbf{Weight learning.}
To learn the weights $\bv{w}$ in Eq.~\eqref{eq:inference}, ground truth summaries for  query-video pairs are required.
Previous methods typically only optimized for relevance~\cite{Liu2015} or used small datasets with limited vocabularies~\cite{Sharghi2016}.
Thus, to be able to train our model, we collected a new dataset with relevance and diversity annotations, which we introduce in the next Section.

If relevance and diversity labels are known, we can estimate the optimal mixing weights of the submodular functions through subgradient descent~\cite{Lin}.
In order to directly optimize for the F1-score used at test time,
we use a locally modular approximation based on the procedure of~\cite{narasimhan2012submodular} and optimize the weights using AdaGrad~\cite{Duchi2011}.

\section{Relevance And Diversity Dataset (RAD)}
\label{sec:rad}
We collected a dataset with query relevance and diversity annotation to let us train and evaluate query-relevant summaries. Our dataset consists of $200$ videos, each of which was retrieved given a different query.

Using Amazon Mechanical Turk (AMT) we first annotate the video frames with query relevance labels, and then partition the frames into  clusters according to visual similarity. These kind of labels were used previously in the MediaEval diverse social images challenge~\cite{Ionescu2015} and enabled evaluation of the automatic methods for creating relevant and diverse summaries.

To select a representative sample of queries and videos for the dataset, we used the following procedure: We take the top YouTube queries between $2008$ and $2016$ from $22$ different categories as seed queries\footnote{https://www.google.com/trends/explore}.\ These queries are typically rather short and generic concepts, so to obtain longer, more realistic queries we use YouTube auto-complete to suggest phrases. Using this approach we collect $200$ queries. Some examples are \textit{brock lesnar vs big show}, \textit{taylor swift out of the woods}, etc.  For each query, we take the top video result with a duration of $2$ to $3$ minutes.

\begin{table*}[t]
\centering
\begin{tabular}{@{}llccclHlll@{}}
	\toprule 
	& & \multicolumn{3}{c}{Settings}                                                              &  & \multicolumn{4}{c}{Metrics}                                        \\ \cmidrule(lr){3-5} \cmidrule(l){7-10} 
	  & Method                                 & Cost        & LSTM         & Quality      &   & HIT@1 VG       & HIT@1 VG or G   & Spear Corr.    & mAP             \\ \midrule
	  & Random                                 & -           & -            & -            &   & 28.2 $\pm$ 1.5 & 57.17 $\pm$ 1.5 & -              & 0.5780          \\ \midrule

	  & Loss of Liu~\etal                     & $l_1$       & $\times$     & $\times$     &   & -              & 68.75           & 0.186         & 0.6308          \\
& Ours: L1                    & $l_1$ & $\times$  & $\times$     &   & -          & 68.09           & 0.209       & 0.6348         \\
& Ours: Huber                    & $l_{huber}$ & $\times$  & $\times$     &   & -          & 68.35           & 0.279         & 0.6446         \\
	  & Loss of Liu~\etal + LSTM               & $l_1$       & $\checkmark$ & $\times$     &   & -         & 70.62           & 0.270          & 0.6507          \\
	  & Ours: Huber + LSTM                     & $l_{huber}$ & $\checkmark$ & $\times$     &   & 36.71          & 72.63           & 0.367          & 0.6685          \\
	  & Ours: Frame quality only $Q_{expli}$ & $l_{huber}$ & $\times$     & $\checkmark$ &   & 31.05          & 65.95           & 0.236          & 0.6315          \\      
	  & Ours: Huber + LSTM + $Q_{impli}$       & $l_{huber}$ & $\checkmark$ & $\checkmark$ &   & -          & 70.76           & 0.371         & 0.6657          \\
	  & Ours: Huber + LSTM + $Q_{expli}$       & $l_{huber}$ & $\checkmark$ & $\checkmark$ &   & 38.86          & \textbf{74.76}  & \textbf{0.376} & \textbf{0.6712} \\

	\bottomrule
\end{tabular}
\caption{Comparison of different model configurations trained on a subset of the Clickture dataset and fine-tuned on our Video Thumbnail dataset (RAD). We report the HIT@1 (fraction of times we select a ``Very Good'' or ``Good'' thumbnail), the Spearman correlation of our model predictions with the true candidate thumbnail scores, and mean average precision.  The Huber+LSTM+$Q_{expli}$ model performs best.}
\label{tab:detailed-expts}
\vspace{-0.3cm}
\end{table*}
To annotate the videos, we set up two consecutive tasks on AMT. All videos are sampled at one frame per second. In the first task, a worker is asked to label each frame with its relevance ~\wrt the given query. Options for answers are ``Very Good'',``Good'', ``Not good'' and ``Trash'', where trash indicates that the frame is both irrelevant and low-quality (\eg blurred, bad contrast, etc.). After annotating the relevance, the worker is asked to distribute the frames into clusters according to their visual similarity. We obtain one clustering per worker, where each clustering consists of mutually exclusive subsets of video frames as clusters. The number of clusters in the clustering is chosen by the worker.
Each video is annotated by $5$ different people and a total of $48$ subjects participated in the annotation.
To ensure high-quality annotations, we defined a qualification task, where we check the results manually to ensure the workers provide good annotations. Only workers who pass this test are allowed to take further assignments.

\subsection{Analysis}
\label{sec:rad_analsis}
We now analyse the two kinds of annotations obtained through this procedure and describe how we merge these annotations into one set of ground truth labels per video.

\bigskip
\noindent
\textbf{Label distributions.}
The distribution of relevance labels is ``Very Good'': $17.55\%$, ``Good'': $57.40\%$, ``Not good'': $12.31\%$  and ``Trash'': $12.72\%$. The minimum, maximum and mean number of clusters per video are $4.9$, $25.2$ and $13.4$ respectively over all videos of RAD.

\bigskip
\noindent
\textbf{Relevance annotation consistency.}
Given the inherent subjectivity of the task, we want to know whether annotators agree with each other about the query relevance of frames.
To do this, we follow previous work~\cite{Isola2011a,GygliICCV13,wang2016event} and compute the Spearman rank correlation ($\rho$) between the relevance scores of different subjects, splitting five annotations of each video into two groups of two and three raters each. We take all split combination to find mean $\rho$ for a video.

Our dataset has an average correlation of $\rho=0.73$ over all videos, where $1$ is a perfect correlation while $0$ would indicate no consistency in the scores. On the related task of event-specific image importance, using five annotators, consistency is only $\rho=0.4$~\cite{wang2016event}. Thus, we can be confident that our relevance labels are of high quality.

\bigskip
\noindent
\textbf{Cluster consistency.}
To the best of our knowledge, we are the first to annotate multiple clusterings per video and look into the consistency of multiple annotators. MediaEval, for example, used multiple relevance labels but only one clustering~\cite{Ionescu2015}. Various ways of measuring the consistency of clusterings exist, \eg Variation of Information, Normalised Mutual Information or the Rand index (See Wagner and Wagner~\cite{Wagner2007} for an excellent overview).
In the following we propose to use Normalised Mutual Information (NMI), an information theoretic measure~\cite{Fred2003} which is the ratio of the mutual information between two clusterings ($I(C,C')$) and the sum of entropies of the clusterings ($H(C) + H(C')$): 

\begin{equation}
NMI(C,C') = \frac{2\cdot I(C,C')}{H(C) + H(C')},
\label{eq:nmi}
\end{equation}
We chose NMI over the more recently proposed Variation of Information (VI)~\cite{meilua2003comparing}, as NMI has a fixed range ($\left[0,1\right]$) while still being closely related to VI (see supplementary material).

Our dataset has a cluster consistency of $0.54$.  Since NMI is $0$ if two clusterings are independent and $1$ iff they are identical, we see that our annotators have a high degree of agreement.

\bigskip
\noindent
\textbf{Ground truth}
For evaluation on the test videos, we create a single ground truth annotation for each video. We merge the five relevance annotations as well as the clustering of each query-video pair. For the final ground truth of relevance prediction, we require the labels be either positive or negative for each video frame. 
We map all ``Very Good'' labels to $1$, ``Good'' labels to $0.5$ and ``Not Good'' and ``Trash'' labels to $0$. We compute the mean of the five relevance annotation labels and label the frame as positive if the mean is $\geq0.5$ and as negative otherwise.

To merge clustering annotations, we calculate NMI between all pairs of clustering and choose the clustering with the highest mean NMI,~\ie the most prototypical cluster. An example of relevance and clustering annotation is provided in Fig.~\ref{fig:rad-summary}.

\section{Configuration testing}
\label{sec:config}
Before comparing our proposed relevance model against state of the art in Sec.~\ref{sec:expts}, we first analyze our model performance using different objectives, cost functions and text representation. 
For evaluation, we use Query-dependent Thumbnail Selection Dataset (QTS) provided by \cite{Liu2015}. The dataset contains $20$ candidate thumbnails for each video, each of which is labeled one of the five: Very Good (VG), Good (G), Fair (F), Bad (B), or Very Bad (VB).  We evaluate on the available $749$ query-video pairs.
To transform the categorical labels to numerical values, we use the same mapping as~\cite{Liu2015}.

\vspace{-0.1cm}
\bigskip
\noindent
\textbf{Evaluation metrics.}
As evaluation metrics, we are using HIT@1 and mean Average Precision (mAP) as reported and defined in Liu~\etal~\cite{Liu2015}, as well as the Spearman's Rank Correlation. 
HIT@1 is computed as the hit ratio for the highest ranked thumbnail.

\vspace{-0.1cm}
\bigskip
\noindent
\textbf{Training dataset.}
For training, we use two datasets: (i) the Bing Clickture dataset~\cite{hua2013clickture} and (ii) the RAD dataset (Sec.~\ref{sec:rad}).
Clickture is a large dataset consisting of queries and retrieved images from Bing Image search.
The annotation is in form of triplets  $(K, Q, C)$ meaning that the image $K$ was clicked $C$ times in the search results of the query $Q$.
This dataset is well suited for training our relevance model, since our task is the retrieval of relevant keyframes from a video, given a text query.
It is, however, from the image and not the video domain.
Thus, we additionally fine-tune the models on the complete RAD dataset consisting of $200$ query-video pairs.
From each query-video pair, we sample an equi number of positive and negative frames to give equal weight to each video.
In total, we use $0.5M$ triplets (as in Sec.~\ref{sec:textrep}) from the Clickture and $14K$ triplets from the RAD for training.

\vspace{-0.15cm}
\bigskip
\noindent
\textbf{Implementation details.} 
We preprocess the images as in ~\cite{simonyan2014very}. We truncate the number of words in the query at $14$, as a tradeoff between the mean and maximum query length in Clickture dataset($5$ and $26$ respectively)~\cite{mueller2016siamese}.
We set the margin parameter $\gamma$ in the loss in Eq.~\eqref{eq:loss} to 1 and the tradeoff parameter $\delta$ for the Huber loss to $1.5$ as in~\cite{Gygli2016}.
The LSTM consists of a hidden layer with $512$ units.
We train the parameters of the LSTM and projection layer $M$ using stochastic gradient descent with adaptive weight updates (AdaGrad)~\cite{Duchi2011}. We add an $l_2$ penalty on the weights, with a $\lambda$ of $10^{-3}$. We train for $20$ epochs using minibatches of $128$ triplets.
\subsection{Tested components}
We discuss three important components of our model next.

\vspace{-0.1cm}
\bigskip
\noindent
\textbf{Objective.}
We compare our proposed training objective to that of Liu~\etal~\cite{Liu2015}. Their model is trained to rank a positive query higher than a negative query given a fixed frame.  In contrast, our method is trained to rank a positive frame higher than a negative frame given a fixed query.

\vspace{-0.1cm}
\bigskip
\noindent
\textbf{Cost function.}
We also investigate the importance of modeling frame quality.
In particular, we compare different cost functions.
(i) We enforce two ranking constraints: one for the quality term and one for the embedding similarity, as in~Eq.\eqref{eq:obj2} ($Q_{expli}$), (ii) We sum the quality and similarity term into one output score, for which we enforce the rank constraint, as in~Eq.\eqref{eq:obj1} ($Q_{impli}$) or (iii) we don't model quality at all.

\vspace{-0.2cm}
\bigskip
\noindent
\textbf{Text representation.}
As mentioned in Sec.~\ref{sec:textrep}, we represent the words of the query using word vectors.
To combine the individual word representations into single vector, we investigate two approaches:
(i) averaging the word embedding vectors and
(ii) using an LSTM model that learns to combine the individual word embeddings.

\subsection{Results}
We show the results of our detailed experiments in Tab.~\ref{tab:detailed-expts}.
They give insights on several important points.

\bigskip
\noindent
\textbf{Text representation.} Modeling queries with an LSTM, rather than averaging the individual word representations, improves performance significantly. This is not surprising, as this model can learn to ignore words that are not visually informative (\eg 2014). 

\begin{table}[t]
\resizebox{1.0\columnwidth}{!}{
\centering
\begin{tabular}{@{}llllll@{}}
\toprule
 &            & \multicolumn{2}{c}{HIT @ 1} &   &         \\ 
 \cmidrule{3-4}

&Method         &VG &  VG or G  & Spear. $\rho$    & mAP         \\ \midrule
\multicolumn{2}{l}{\textsc {Queries}}   \\
& Liu~\etal\cite{Liu2015} & \textbf{40.625} & 73.83 & 0.122& 0.629\\

\multicolumn{2}{l}{\textsc {Titles}}\\
& QAR without $Q_{expli}$ \comment{Ours: CNN-LSTM}       &  36.71& 72.63  & 0.367         & 0.6685 \\
& \textbf{QAR (Ours)} \comment{Ours: CNN-LSTM +$Q_{expli}$}      &  38.86 & \textbf{74.76}  & \textbf{0.376}         & \textbf{0.6712} \\
\bottomrule
\end{tabular}
}
\caption{Comparison of thumbnail selection performance against the state of the art, on the QTS evaluation dataset. Note that~\cite{Liu2015} uses queries for their method which are not publicly available (see text).
}
\label{tab:msr-relevance}
\vspace{-0.2cm}
\end{table}
\begin{figure}[t]
\adjustbox{trim={.0\width} {.04\height} {0.05\width} {.06\height},clip}{\includegraphics[width=0.95\linewidth]{\EmbeddingRoot/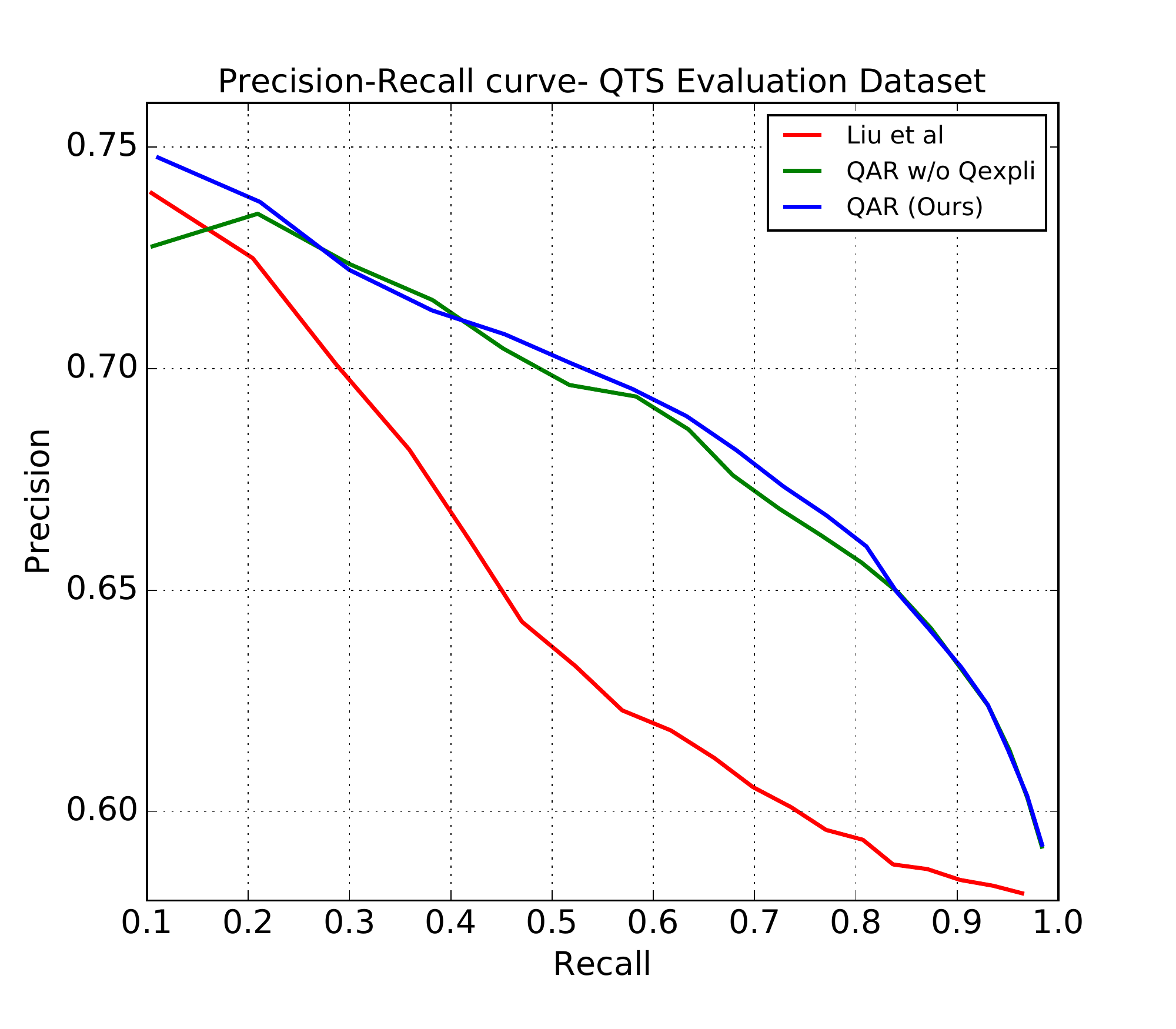}}
\vspace{-0.4cm}
\caption{Precision-Recall curve of QTS Evaluation dataset for different methods.}
\label{fig:PRcurveQTS}
\vspace{-0.3cm} 
\end{figure}
\bigskip
\noindent
\textbf{Objective and Cost function.} 
The analysis shows that training with our objective leads to better performance compared to using the objective of Liu~\etal~\cite{Liu2015}.
This can be explained with the properties of videos, which typically contain many frames that are low-quality or not visually informative~\cite{song2016click}.
Thus, formulating the thumbnail task in a way that the model can learn about these quality aspects is beneficial.
Using the appropriate triplets for training boosts performance substantially (correlation with the loss of Liu~\etal~\cite{Liu2015} + LSTM: $0.270$, Ours: Huber + LSTM $0.367$).
When including a quality term in the model, performance improves further, where an explicit loss performs slightly better (Ours: Huber + LSTM + $Q_{expli}$ in Tab.~\ref{tab:detailed-expts}). 

Somewhat surprisingly, modeling quality alone already outperforms Liu~\etal~\cite{Liu2015} in terms of mAP, despite not using any textual information. 
Quality adds a significant boost to performance in the video domain. Interestingly, this is different in the image domain, due to the difference in quality statistics.
Images returned by a search engine are mostly of good quality, thus explicitly accounting for it does not improve performance (see supplementary material).
\begin{table}[t]
\resizebox{1.0\columnwidth}{!}{
\centering
\begin{tabular}{@{}lllll@{}}
\toprule
& Method                   & HIT@1   & Spear. $\rho$  & mAP            \\ \midrule
\multicolumn{2}{l}{\textsc {No textual input}}\\
& Random                   & 66.6 $\pm$ 3.5 &     $0.0$         & 0.674          \\
& Video2GIF~\cite{Gygli2016}  & 67.0 & \textbf{0.167}   & 0.708          \\
& Ours: Frame quality $Q_{expli}$  & \textbf{69.0} & 0.135   & \textbf{0.749}          \\
\multicolumn{2}{l}{\textsc {Titles}}\\
& Liu~\etal~\cite{Liu2015} +LSTM          & 70.0 &  0.134   & 0.731         \\
& QAR without $Q_{expli}$  &  70.0 &  0.182 & 0.743 \\
& \textbf{QAR (Ours)} \comment{Ours: CNN-LSTM + $Q_{expli}$} &  \textbf{71.0} &  \textbf{0.221} & \textbf{0.760} \\
\multicolumn{2}{l}{\textsc {Queries}}\\
& Liu~\etal~\cite{Liu2015} +LSTM        & 72.0           & 0.204         & 0.730         \\
& QAR without $Q_{expli}$    &  \textbf{76.0} &  \textbf{0.268} & 0.752 \\
& \textbf{QAR (Ours)} \comment{Ours: CNN-LSTM + $Q_{expli}$}   &  72.0 &  0.264 & \textbf{0.769} \\
\bottomrule
\end{tabular}
}
\caption{Performance of our relevance models on the RAD dataset in comparison with previous methods.}
\label{tab:rad-relevance}
\vspace{-0.36cm}
\end{table}
\begin{figure}[t]
\adjustbox{trim={.0\width} {.04\height} {0.05\width} {.06\height},clip}{\includegraphics[width=0.95\linewidth]{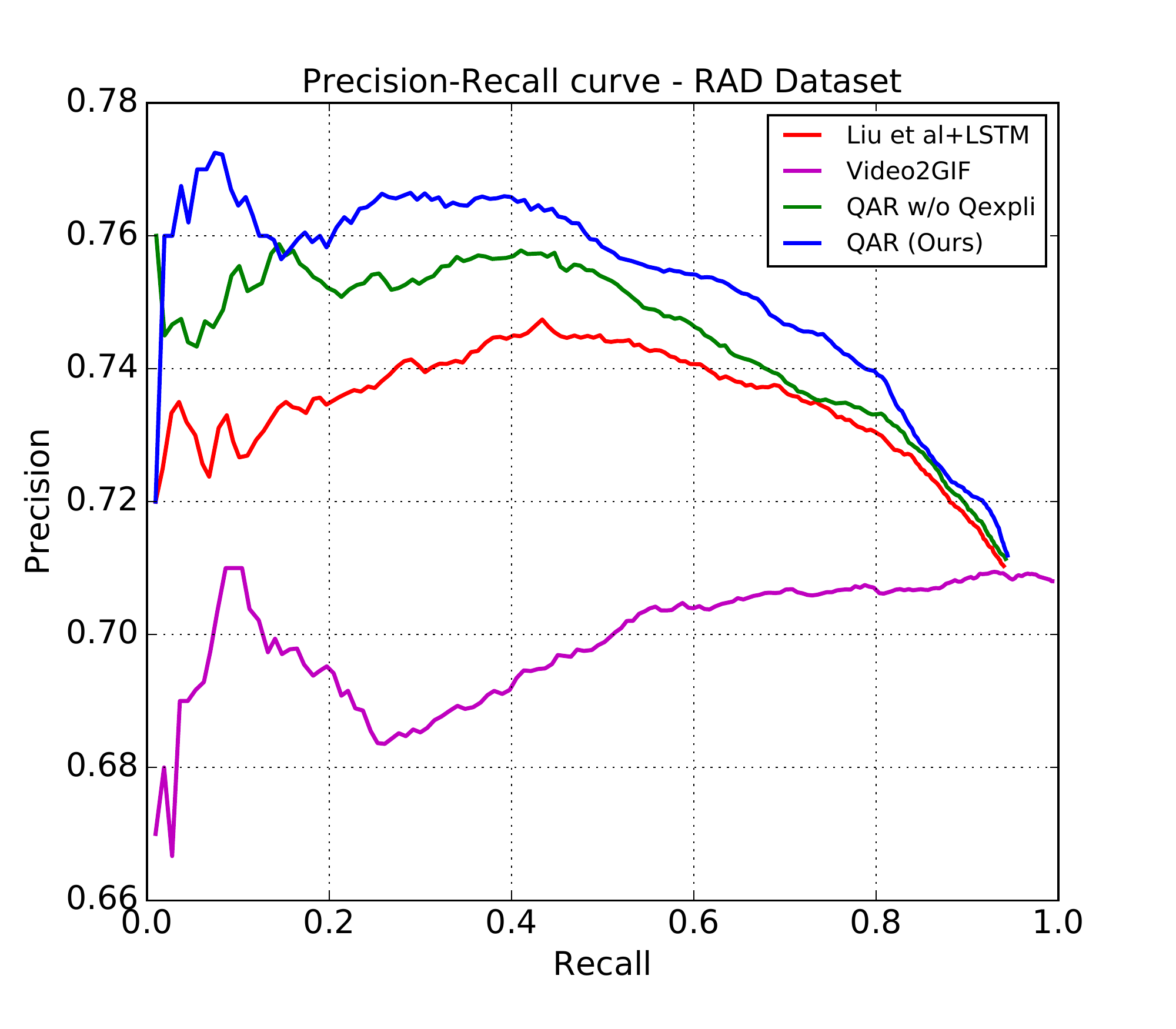}}
\vspace{-0.4cm}
\caption{Recall-Precision curve of the RAD testet for different methods. Our method ({\color{blue}Blue}) performs high in terms of mAP.}
\label{fig:PRcurve-RAD}
\vspace{-0.35cm}
\end{figure}

To conclude, we see that the better alignment of the objective to the keyframe retrieval task, the addition of an LSTM and modeling quality of the thumbnails improves performance.
Together, they provide an substantial improvement compared to Liu~\etal's model.
Our method achieves an absolute improvement of $6.01$\% in HIT@1,  $4.04$\% in mAP, and an improvement in correlation from $0.186$ to $0.376$.
These gains are even more significant when we consider the possible ranges of these metrics.~\eg for Spearman correlation, human agreement is at $0.73$ on the RAD dataset (\cf Sec.~\ref{sec:rad_analsis}), thus providing an upper bound. Similarly, HIT@1 and mAP have small effective ranges given their high scores for a random model.

\section{Experiments}
\label{sec:expts}

In the previous section, we have determined that our objective, embedding queries with an LSTM and explicitly modelling quality performs best.
We call this model QAR (Quality-Aware Relevance) in the following and compare against state-of-the-art(s-o-a) models on the QTS and RAD datasets.
We also evaluate the full summarization model on RAD. For these experiments, we split RAD into $100$ videos for training, $50$ for validation and $50$ for testing.

\bigskip
\noindent
\textbf{Evaluation metrics.}
For relevance we use the same metrics as in Sec.~\ref{sec:config}.
To evaluate video summaries on RAD, we additionally use F1 scores. The F1 score is the harmonic mean of precision of relevance prediction and cluster recall~\cite{Ionescu2015}. It is high, if a method selects relevant frames from diverse clusters.

\subsection{Evaluating the Relevance Model}
We evaluate our model (QAR) and compare it to Liu~\etal~\cite{Liu2015} and Video2GIF \cite{Gygli2016}.

\begin{figure}[t]
    \centering
    {\small Liu~\etal~\cite{Liu2015}}\hspace{15pt} {\small Video2GIF~\cite{Gygli2016}}\hspace{35pt} {\small Ours\hspace{80pt}}\\
    \begin{turn}{90}{\scriptsize Ariana Grande}\end{turn}\begin{turn}{90}{\scriptsize dangerous}\end{turn}
    \hspace{-0.15cm}
    \adjustbox{trim={.1\width} {.15\height} {0.1\width} {.15\height},clip}{\includegraphics[width=0.18\textwidth]{\EmbeddingRoot/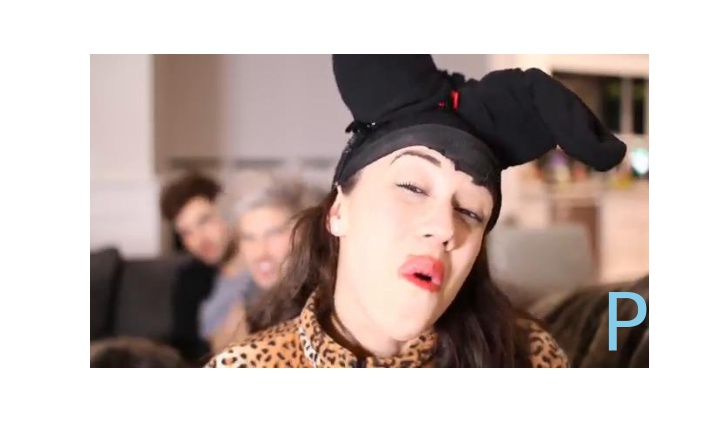}}
    \adjustbox{trim={.1\width} {.15\height} {0.1\width} {.15\height},clip}
    {\includegraphics[width=0.18\textwidth]{\EmbeddingRoot/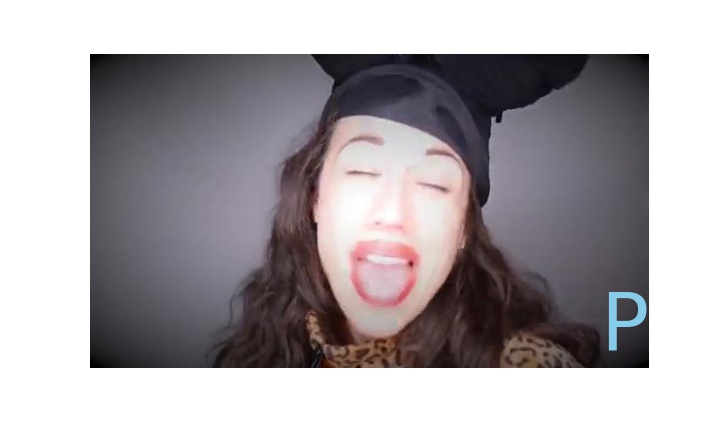}}
    \adjustbox{trim={.1\width} {.15\height} {0.1\width} {.15\height},clip}
    {\includegraphics[width=0.18\textwidth]{\EmbeddingRoot/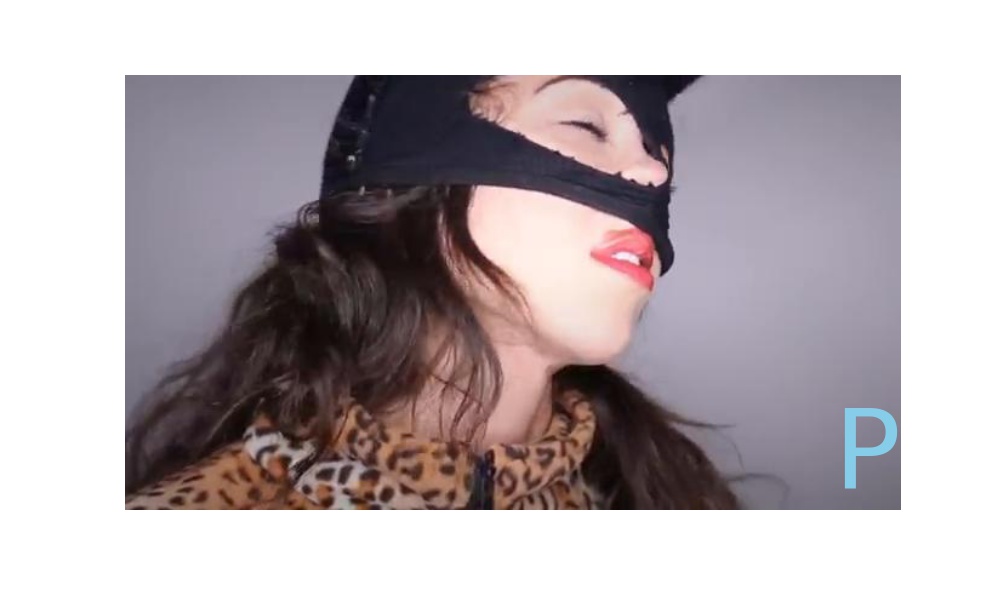}} \\
    \vspace{0.08cm}
    \begin{turn}{90}{\scriptsize Anaconda }\end{turn}\begin{turn}{90}{\scriptsize vs lion}\end{turn}
    \adjustbox{trim={.1\width} {.15\height} {0.12\width} {.15\height},clip}{\includegraphics[width=0.18\textwidth]{\EmbeddingRoot/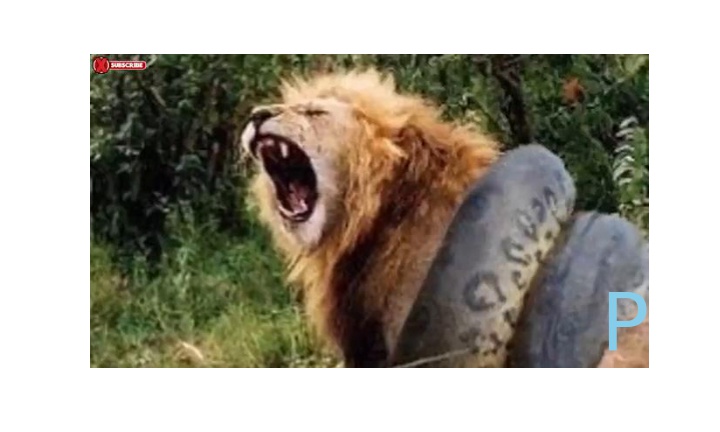}  }
    \adjustbox{trim={.12\width} {.15\height} {0.1\width} {.15\height},clip}
{    \includegraphics[width=0.18\textwidth]{\EmbeddingRoot/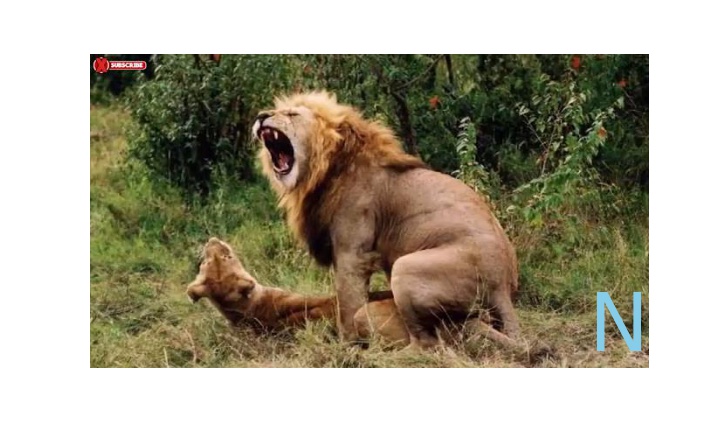}}
\adjustbox{trim={.1\width} {.15\height} {0.1\width} {.15\height},clip}
    {\includegraphics[width=0.18\textwidth]{\EmbeddingRoot/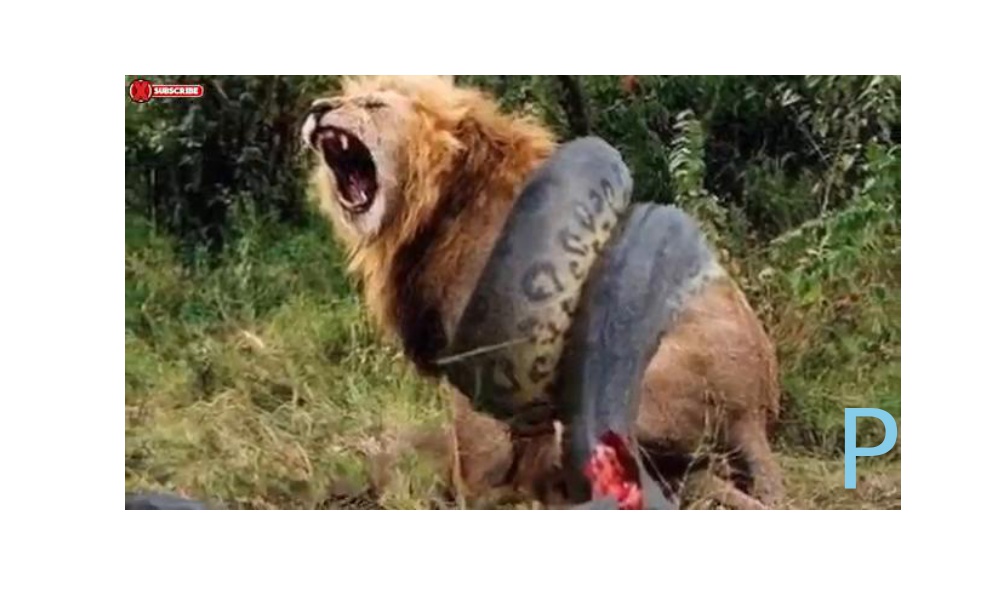}} \\
    \vspace{0.08cm}
    \begin{turn}{90}{\scriptsize Truck tug}\end{turn}\begin{turn}{90}{\scriptsize of war}\end{turn}
    \hspace{-0.15cm}
    \adjustbox{trim={.1\width} {.15\height} {0.1\width} {.15\height},clip}{\includegraphics[width=0.18\textwidth]{\EmbeddingRoot/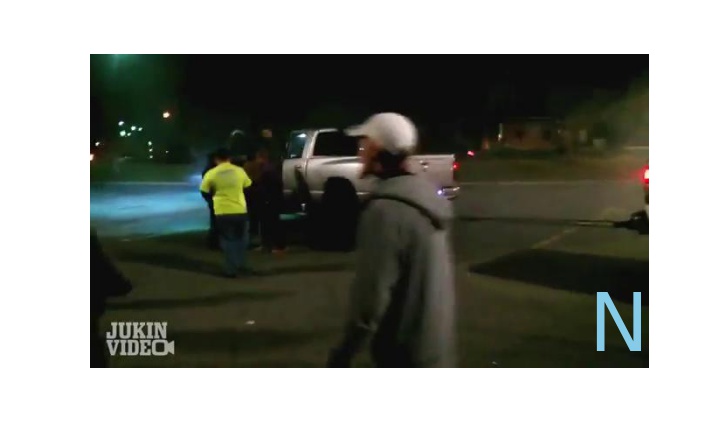}}
    \adjustbox{trim={.1\width} {.15\height} {0.1\width} {.15\height},clip}
    {\includegraphics[width=0.18\textwidth]{\EmbeddingRoot/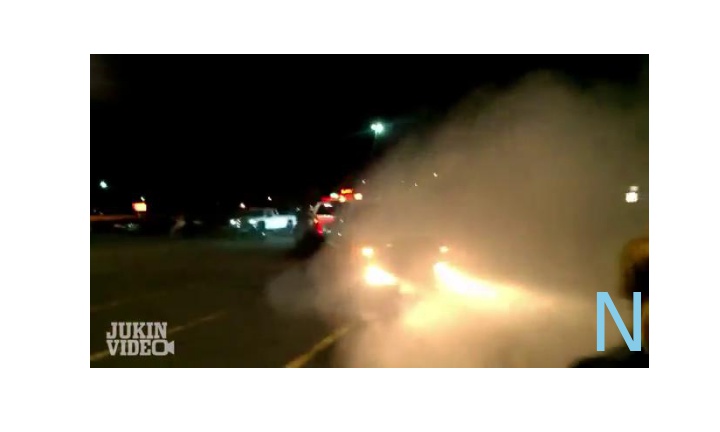}}
    \adjustbox{trim={.1\width} {.15\height} {0.1\width} {.15\height},clip}
    {\includegraphics[width=0.18\textwidth]{\EmbeddingRoot/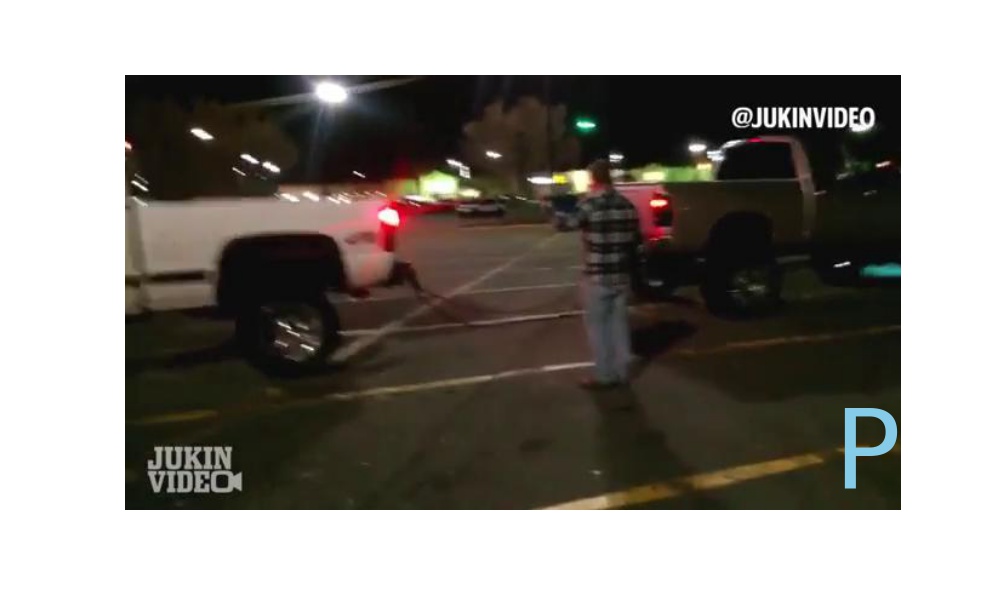}} \\
    \vspace{0.08cm}
    \begin{turn}{90}{\scriptsize Baby alive}\end{turn}
    \hspace{0.05cm}
    \adjustbox{trim={.1\width} {.15\height} {0.1\width} {.15\height},clip}{\includegraphics[width=0.18\textwidth]{\EmbeddingRoot/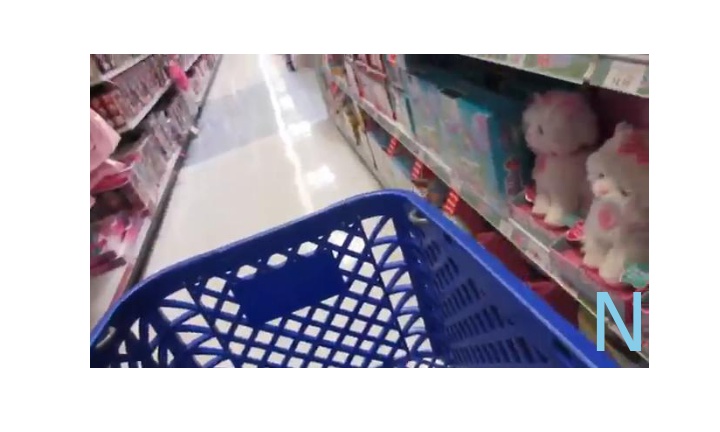}}
    \adjustbox{trim={.1\width} {.15\height} {0.1\width} {.15\height},clip}
    {\includegraphics[width=0.18\textwidth]{\EmbeddingRoot/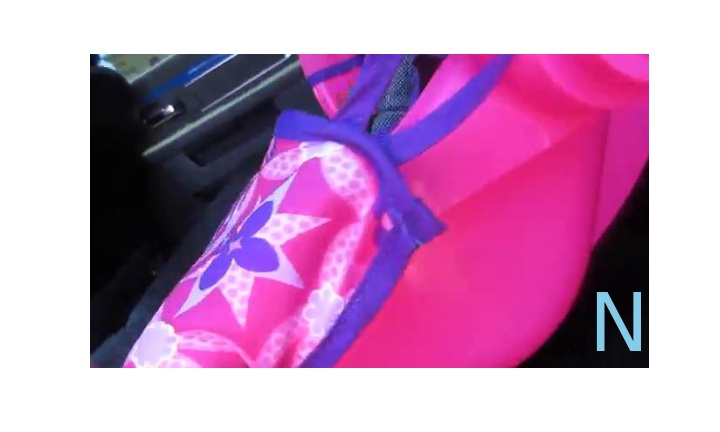}}
    \adjustbox{trim={.1\width} {.15\height} {0.1\width} {.15\height},clip}
    {\includegraphics[width=0.18\textwidth]{\EmbeddingRoot/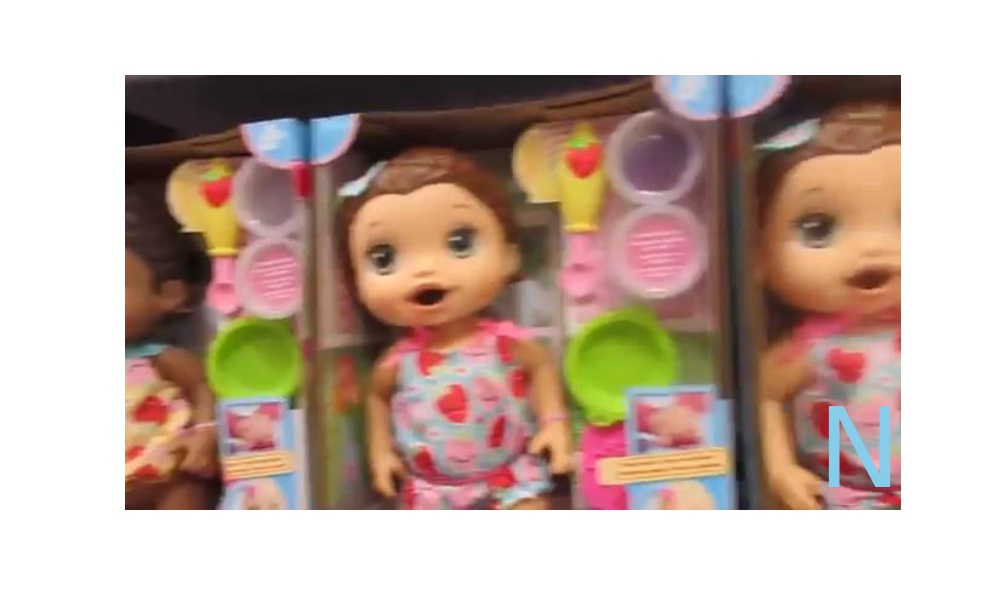}}
    \caption{Qualitative Results of top ranked keyframes on RAD. a) Liu~\etal b) Video2GIF c) Our model (from left). Video titles are shown on the left. 
Ground truth relevance labels are shown in {\color{blue}{Blue}}. P$=$Positive, N$=$Negative.} 
\label{fig:relevancecomparison}
\vspace{-0.1cm}
\end{figure} 

\bigskip
\noindent
\textbf{Query-dependent Thumbnail Selection Dataset (QTS)~\cite{Liu2015}}

We compare against the s-o-a on the QTS evaluation dataset in Tab.~\ref{tab:msr-relevance}.
We report the performance of Liu~\etal~\cite{Liu2015} from their paper. 
Note, however, that the results are not directly comparable, as they use query-video pairs for predicting relevance, while only the titles are shared publicly.
Thus, we use the titles instead, which is an important difference.
Relevance is annotated with respect to the queries, which often differ from the video titles.
We compare the re-implementation of \cite{Liu2015} using titles in detail in Tab.~\ref{tab:detailed-expts}.

Encouragingly, our model performs well even when just using the titles and outperforms them on most metrics. It improves mAP by $4.22$\% over ~\cite{Liu2015} and correlation by a margin of 0.254 (\cf Table~\ref{tab:msr-relevance}). Figure~\ref{fig:PRcurveQTS} shows the precision-recall curve for the experiment. As can be seen QAR outperforms~\cite{Liu2015} for all recall ratios.
To better understand the effects of using titles or queries, we quantify the value of the two on the RAD dataset below.


\begin{figure*}[t]
\centering
    \textsc{Query}: Hairstyles for Men\\
\begin{turn}{90}{\small \hspace{0.3cm} Hecate~\cite{song2016click}}\end{turn}
    \adjustbox{trim={.125\width} {.255\height} {0.1\width} {.23\height},clip}
    {\includegraphics[width=1.08\textwidth]{\EmbeddingRoot/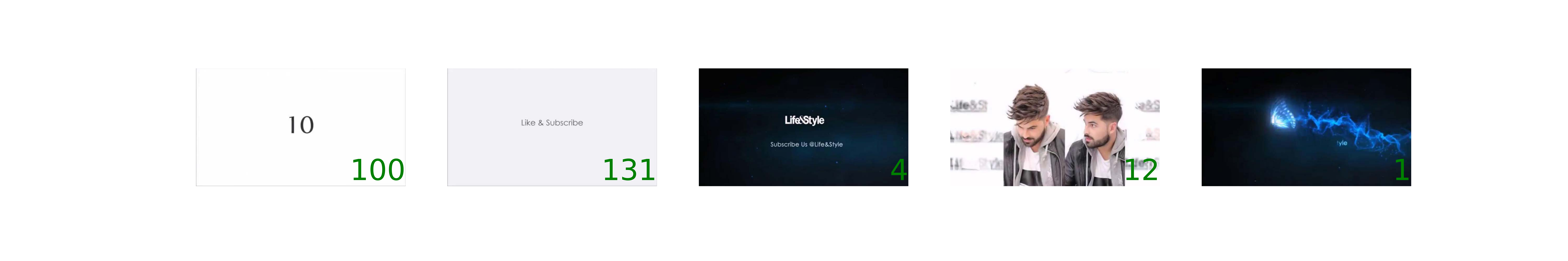}} \\
\begin{turn}{90}{\small \hspace{0.3cm} MMR~\cite{carbonell1998use}}\end{turn}
    \adjustbox{trim={.125\width} {.255\height} {0.1\width} {.23\height},clip}
    {\includegraphics[width=1.08\textwidth]{\EmbeddingRoot/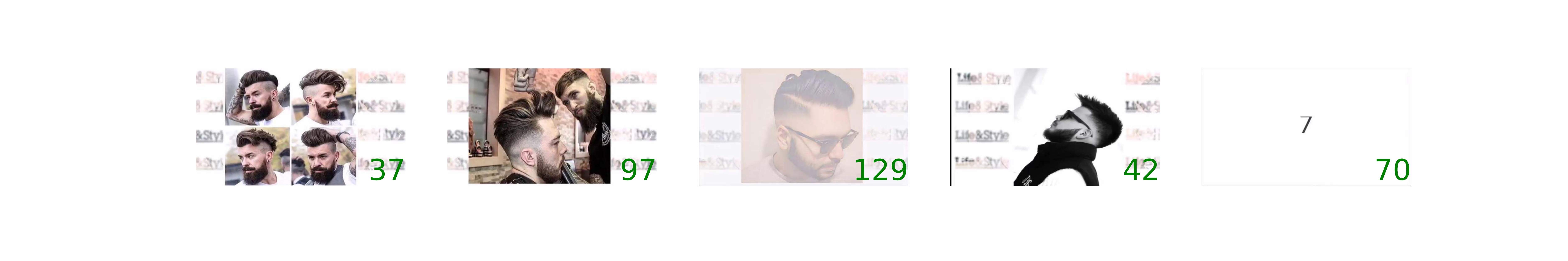}} \\
\begin{turn}{90}{\small \hspace{0.3cm} Similarity}\end{turn}
    \adjustbox{trim={.125\width} {.255\height} {0.1\width} {.23\height},clip}
    {\includegraphics[width=1.08\textwidth]{\EmbeddingRoot/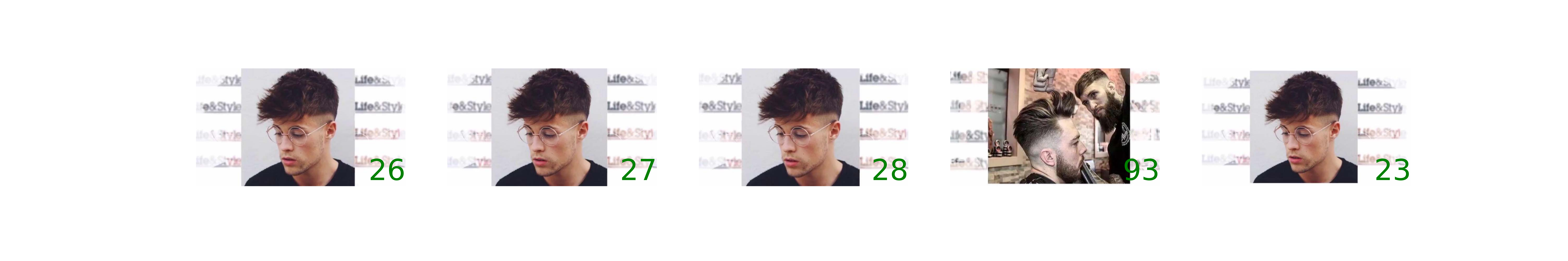}} \\
    \begin{turn}{90}{\small \hspace{0.7cm} Ours}\end{turn}
    \adjustbox{trim={.125\width} {.22\height} {0.1\width} {.23\height},clip}
    {\includegraphics[width=1.08\textwidth]{\EmbeddingRoot/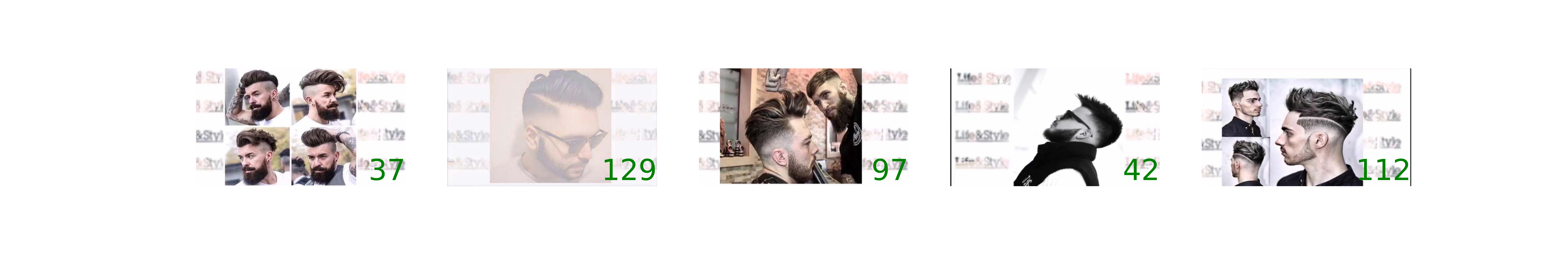}} \\
   \adjustbox{trim={.06\width} {0.20\height} {0.05\width} {0.58\height},clip}
   {\includegraphics[width=1.08\textwidth]{\EmbeddingRoot/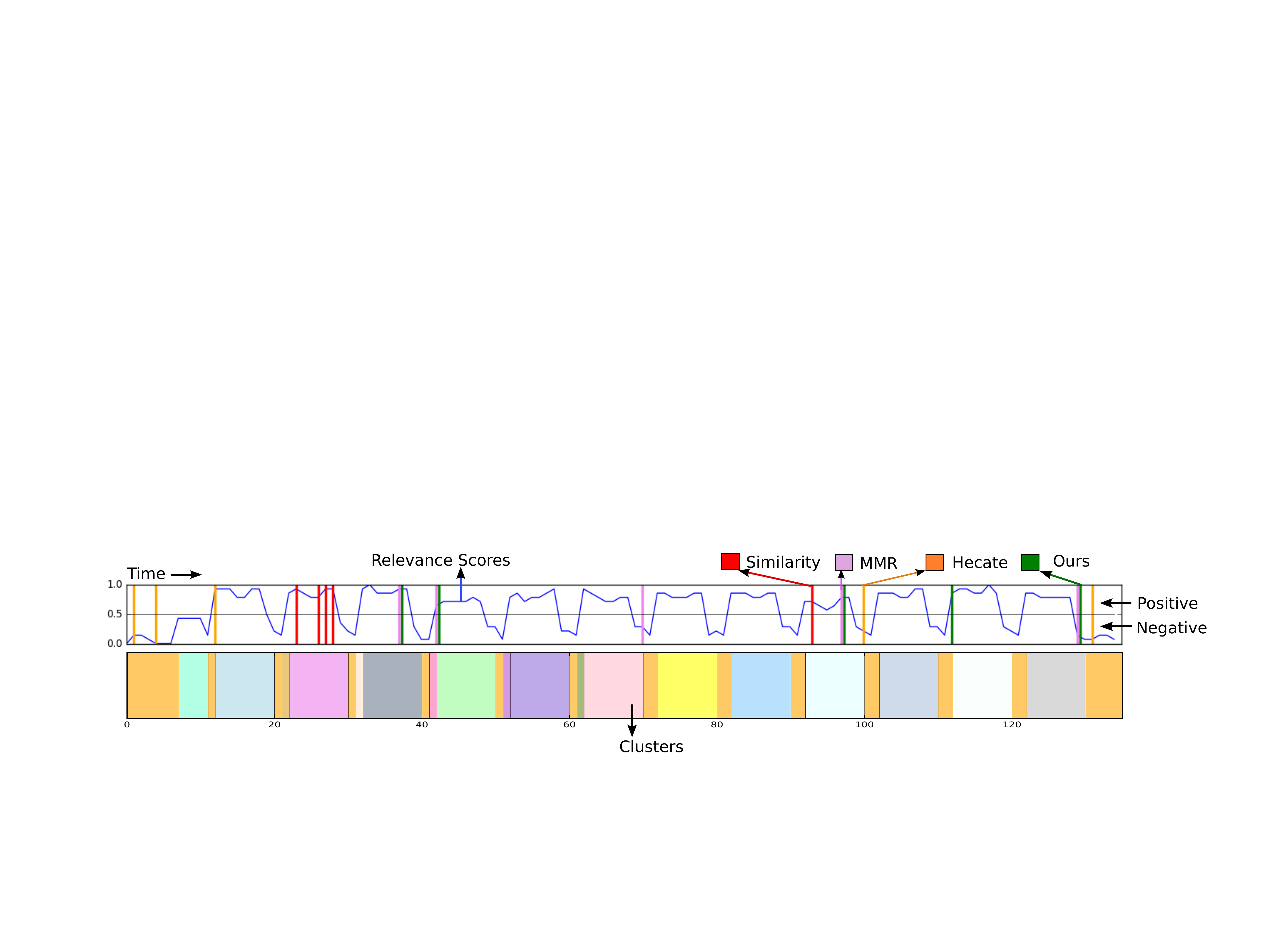}}\\
\vspace{-0.4cm}
\caption{We show video summaries created by Hecate~\cite{song2016click}, MMR~\cite{carbonell1998use}, our similarity model and our full summarization approach. The {\color{green} Green} number on the images depicts the frame number. We plot the ground truth relevance scores, marking the selected frames for the shown methods, and cluster annotations over the video in the bottom two rows. For cluster annotation, each color represents a unique cluster. Additional examples are provided in supplementary. }
\label{fig:rad-summary}
\vspace{-0.2cm}
\end{figure*}
\bigskip
\noindent
\textbf{Our dataset (RAD)}
We also evaluate our model on the RAD test set (Tab.~\ref{tab:rad-relevance}).
QAR (ours) significantly outperforms the previous s-o-a of \cite{Liu2015,Gygli2016}, even when augmenting Liu~\etal\cite{Liu2015} with an LSTM.
QAR improves mAP by $2.9$\% when using \textsc{Titles} and $3.9$\% when using \textsc{Queries} over our implementation of Liu~\etal~\cite{Liu2015}+LSTM.

We also see that modeling quality leads to significant gains in terms of mAP when using \textsc{Titles} or \textsc{Queries} ($+1.7\%$ in both cases).
HIT@1 for query relevance, however, is lower when including quality.
We believe that the reason for this is that when the query is given, the textual-visual similarity is a more reliable signal to determine the single best keyframe.
While including quality improves the overall ranking on mAP, it is solely based on appearance and thus seems to inhibit the fine-grained ranking results at low recall(Fig.~\ref{fig:PRcurve-RAD}).
However, when only the title is used, the frame quality becomes a stronger predictor for thumbnail selection and improves performance on all metrics.
We present some qualitative results of different methods for relevance prediction in Fig.~\ref{fig:relevancecomparison}. 

\begin{table}[t]
\resizebox{1.0\columnwidth}{!}{
\centering
\begin{tabular}{@{}llllccc@{}}
\toprule
 \multicolumn{4}{c}{Method}                & $<PR> $  &  $ <CR>$  & $<F1>$    \\ \midrule

Similarity & Diversity & Quality & Repr &  & &\\
  $-$  & $-$ &   $-$ & $\checkmark$ &0.654 & 0.817& 0.672\\
   $-$  & $-$ &  $\checkmark$ & $-$ &0.671 &0.542&0.522\\
      $-$  &$\checkmark$ & $-$ &   $-$ & 0.575 & 0.808& 0.629\\
 $\checkmark$ & $-$ & $-$ & $-$ & 0.763 & 0.550 &0.578 \\
  $\checkmark$ & $-$ & $\checkmark$ & $-$ & \textbf{0.775} & 0.563&0.594 \\
 \midrule

    \multicolumn{4}{c}{MMR~\cite{carbonell1998use}}     & & & \\
    $\checkmark$ (33\%)& $\checkmark$ (66\%) & $-$ & $-$ & 0.692 &\textbf{0.825}& 0.716 \\ 
    \midrule
    
\multicolumn{4}{c}{Hecate~\cite{song2016click}}  & 0.708& 0.787 & 0.713\\
    \midrule
  
\multicolumn{4}{c}{\textbf{Ours}} & & &\\

$\checkmark$ (45\%) & $\checkmark$ (43\%) & $\checkmark$ (2\%) & $\checkmark$ (10\%) & 0.704 & \textbf{0.825} & \textbf{0.721}\\ 
\midrule
\multicolumn{4}{c}{Upper bound}  & 0.938& 0.925 & 0.928\\
\bottomrule
\end{tabular}
}
\caption{Performance of summarization methods on the RAD dataset.\ \textit{Repr} means Representativeness.\ $\checkmark$ and $-$ depict whether an objective was used or not.\ MMR and ours learn their corresponding weights. Percentage in parentheses the normalized learnt weights. Upper bound refers to the best possible performance, obtained using the ground truth annotations of RAD.}
\vspace{-0.5cm}
\label{tab:rad-summary}
\end{table}

\subsection{Evaluating the Summarization Model}
As mentioned in Sec.~\ref{sec:summarization}, we use four objectives for our summarization model. Referring to Tab.~\ref{tab:rad-summary}, we use QAR model to get Similarity and Quality scores while Diversity and Representativeness scores are obtained as described in Sec.~\ref{sec:summarization}. We compare the performance of our full model with each individual objective, a baseline based on Maximal Marginal Relevance (MMR)~\cite{carbonell1998use} and Hecate~\cite{song2016click}. MMR greedily builds a set that maximises the weighted sum of two terms: (i) The similarity of the selected elements to a query and (ii) The dissimilarity to previously selected elements. To estimate the similarity to the query we use our own model (QAR without $Q_{expli}$) and for dissimilarity the diversity as defined in Sec.~\ref{sec:summarization}.
Finally, we compare it to Hecate, recently introduced in~\cite{song2016click}. Hecate estimates frame quality using the stillness of the frame and selects representative and diverse thumbnails by clustering the video with k-means and selecting the highest quality frame from the k largest clusters.

\vspace{-0.15cm}
\bigskip
\noindent
\textbf{Results}
Quantitative results are shown in Tab.~\ref{tab:rad-summary}, while Fig.~\ref{fig:rad-summary} shows qualitative results.
As can be seen, combining all objectives with our model works best. 
It outperforms all single objectives, as well as the MMR~\cite{carbonell1998use} baseline, even though MMR also uses our well-performing similarity estimation.
Similarity alone has the highest precision, but tends to pick frames that are visually similar (\cf Fig.~\ref{fig:rad-summary}), thus resulting in low cluster recall.
Diversification objectives (diversity and representativeness) have a high cluster recall, but the frames are less relevant.
Somewhat surprisingly, Hecate~\cite{song2016click} is a relatively strong baseline. In particular, it performs well in terms of relevance, despite using a simple quality score. This further highlights the importance of quality for the thumbnail selection task.
It also indicates that the used VGG-19 architecture might be suboptimal for predicting quality. CNNs for classification use small input resolutions, thus making it difficult to predict quality aspects such as blur. Finding better architectures for that task is actively researched,~\eg~\cite{lu2015deep,mai2016composition}, and might be used to improve our method.

When analysing the learned weights (\cf Tab.~\ref{tab:rad-summary}) we find that the similarity prediction is the most important objective, which matches our expectations.
Quality gets a lower, but non-zero weight, thus showing that it provides information that is complementary to query-similarity.
Thus, it helps predicting the relevance of a frame.
The reader should however be aware that differences in the variance of the objectives can affect the weights learned. Thus, they should  be taken with a grain of salt and only be considered tendencies.

\section{Conclusion}
We introduced a new method for query-adaptive video summarization. At its core lies a textual-visual embedding, which lets us select frames relevant to a query. In contrast to earlier works, such as~\cite{zeng2016semantic,Sharghi2016}, this model allows us to handle unconstrained queries and even full sentences.
We proposed and empirically evaluated different improvements over~\cite{Liu2015}, for learning a relevance model.
Our empirical evaluation showed that a better training objective, a more sophisticated text model, and explicitly modelling quality leads to significant performance gains.
In particular, we showed that quality plays an important role in the absence of high-quality relevance information, such as queries, \ie when only the title can be used.
Finally, we introduced a new dataset for thumbnail selection which comes with query-relevance labels and a grouping of the frames according to visual and semantic similarity.
On this data, we tested our full summarization framework and showed that it compares favourably to strong baselines such as MMR~\cite{carbonell1998use} and~\cite{song2016click}.
We hope that our new dataset will spur further research in query adaptive video summarization.

\vspace{-0.1cm}
\section{Acknowledgements}
 This work has been supported by Toyota via the project TRACE-Zurich.  We also acknowledge the support by the CHIST-ERA project MUSTER. MG  was  supported  by  the European Research Council under the project VarCity (\#273940).

\bibliographystyle{ACM-Reference-Format}
\bibliography{video_sum,ref}

\end{document}